\documentclass[11pt,a4paper]{article}
\usepackage[hyperref]{emnlp2020}
\usepackage{times}
\usepackage{latexsym}
\usepackage{multirow}

\usepackage{microtype}
\usepackage{graphicx}
\usepackage{colortbl}
\aclfinalcopy % Uncomment this line for the final submission
\usepackage{amsfonts}
\usepackage{amssymb}

\newcommand{\STAB}[1]{\begin{tabular}{@{}c@{}}#1\end{tabular}}

\renewcommand{\vec}[1]{\mathbf{#1}}

\newcommand{\cg}[0]{\cellcolor{gray!10}}

\title{Discontinuous Constituent Parsing as Sequence Labeling}

  \author{
  David Vilares and Carlos G\'{o}mez-Rodr\'{i}guez \\
  Universidade da Coru\~{n}a, CITIC \\
  Departamento de Ciencias de la Computación y Tecnologías de la Información \\
  Campus de Elvi\~{n}a s/n, 15071 \\ A Coru\~{n}a, Spain \\
  {\tt \{david.vilares,carlos.gomez\}@udc.es} 
  \\}

\date{}

\begin{document}

\maketitle

\begin{abstract}
    This paper reduces discontinuous parsing to sequence labeling. It first shows that existing reductions for constituent parsing as labeling
    do not support discontinuities. Second, it fills this gap and proposes to encode tree discontinuities as nearly ordered permutations of the input sequence. Third, it studies whether such discontinuous representations are learnable. The experiments show that despite the architectural simplicity, under the right representation, the models are fast and accurate.\footnote{\url{https://github.com/aghie/disco2labels}}

\end{abstract}

\section{Introduction}

\begin{figure}[tbp]
\centering
\includegraphics[width=0.80\columnwidth]{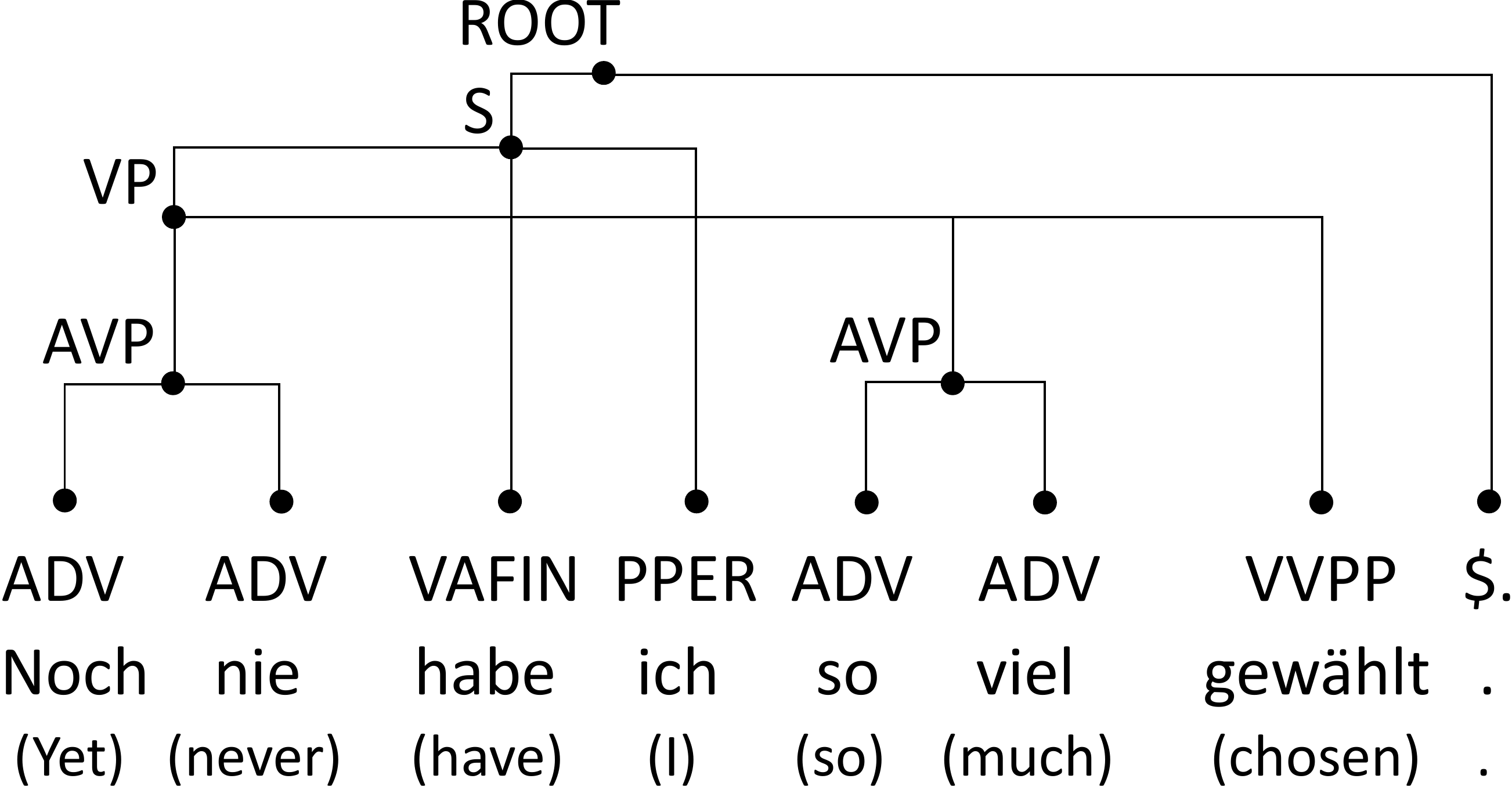}
\caption{\label{f-negra-discontinuous-tree} An example of a German sentence exhibiting discontinuous structures, extracted from the NEGRA treebank \cite{skut1997annotation}. A valid English translation is: \emph{`Never before I have chosen so much.'}}
\end{figure}

Discontinuous constituent parsing studies how to generate phrase-structure trees of sentences coming from \emph{non-configurational} languages \cite{johnson1985parsing}, where non-consecutive tokens can be part of the same grammatical function (e.g. non-consecutive terms belonging to the same verb phrase). Figure \ref{f-negra-discontinuous-tree} shows a German sentence exhibiting this phenomenon. Discontinuities happen in languages that exhibit free word order such as German or Guugu Yimidhirr \cite{haviland1979guugu,johnson1985parsing}, but also in those with high rigidity, e.g. English, whose grammar allows certain discontinuous expressions, such as wh-movement or extraposition \cite{evang2011plcfrs}. This makes discontinuous parsing a core computational linguistics problem that affects a wide spectrum of languages. 

There are different paradigms for discontinuous phrase-structure parsing, such as chart-based parsers \cite{maier2010direct, corro2020span}, transition-based algorithms \cite{coavoux2017incremental,coavoux2019discontinuous} or reductions to a problem of a different nature, such as dependency parsing \cite{hall2008dependency,fernandez2015parsing}. However, many of these approaches come either at a high complexity or low speed,
while others give up significant performance to achieve an acceptable latency \cite{maier2015discontinuous}.

Related to these research aspects, this work explores the feasibility of discontinuous parsing under the sequence labeling paradigm, inspired by \newcite{gomez-rodriguez-vilares-2018-constituent}'s work on fast and simple \emph{continuous} constituent parsing.
We will focus on tackling the limitations of their encoding functions when it comes to analyzing discontinuous structures, and include an empirical comparison against existing parsers.

\paragraph{Contribution} (i) The first contribution is theoretical: to reduce constituent parsing of free word order languages to a sequence labeling problem.
This is done by encoding the order of the sentence as (nearly ordered) permutations. We present various ways of doing so, which can be naturally combined with the labels produced by existing reductions for \emph{continuous} constituent parsing. (ii) The second contribution is a practical one: to show how these representations can be learned by neural transducers. We also shed light on whether general-purpose architectures for NLP tasks \cite{devlin-etal-2019-bert,sanh2019distilbert} can effectively parse free word order languages, and be used as an alternative to \emph{ad-hoc} algorithms and architectures for discontinuous constituent parsing.

\section{Related work}\label{section-related-work}

Discontinuous phrase-structure trees can be derived by expressive formalisms such as \emph{Multiple Context Free Grammmars} \cite{seki1991multiple} (MCFGs) or \emph{Linear Context-Free Rewriting Systems} (LCFRS) \cite{vijay1987characterizing}. MCFGs and LCFRS are essentially an extension of \emph{Context-Free Grammars} (CFGs) such that non-terminals can link to non-consecutive spans. Traditionally, chart-based parsers relying on this paradigm commonly suffer from high complexity \cite{evang2011plcfrs,maier2010discontinuity,maier2010direct}. 
Let $k$ be the block degree, i.e. the number of non-consecutive spans than can be attached to a single non-terminal; the complexity of applying CYK (after binarizing the grammar) would be $\mathcal{O}(n^{3k})$ \citep{seki1991multiple}, which can be improved to $\mathcal{O}(n^{2k+2})$ if the parser is restricted to well-nested LCFRS \citep{gomez-rodriguez-etal-2010-efficient}, and \newcite{maier2015discontinuous} discusses how for a standard discontinuous treebank, $k\approx3$ (in contrast to $k= 1$ in CFGs).
Recently, \newcite{corro2020span} presents a chart-based parser for $k=2$ that can run in $\mathcal{O}(n^{3})$, which is equivalent to the running time of a continuous chart parser, while covering 98\% of the discontinuities. Also recently, \newcite{stanojevic2020span} present an LCFRS parser with $k=2$ that runs in $\mathcal{O}(ln^4+n^6)$ worst-case time, where $l$ is the number of unique non-terminal symbols, but in practice they show that the empirical running time is among the best chart-based parsers.

Differently, it is possible to rely on the idea that discontinuities are inherently related to the \emph{location} of the token in the sentence. In this sense, it is possible to reorder the tokens while still obtaining a grammatical sentence that could be parsed by a continuous algorithm.  This is usually achieved with transition-based parsing algorithms and the \texttt{swap} transition \cite{nivre2009non} which switches the topmost elements in the stack. For instance, \newcite{versley2014incorporating} uses this transition to adapt an easy-first strategy \cite{goldberg2010efficient} for dependency parsing to discontinuous constituent parsing. In a similar vein, \newcite{maier2015discontinuous} builds on top of a fast continuous shift-reduce constituent parser \cite{zhu2013fast}, and incorporates both standard and bundled swap transitions in order to analyze discontinuous constituents.  \citeauthor{maier2015discontinuous}'s system produces derivations of up to a length of $n^2-n+1$ given a sentence of length $n$. More efficiently, \newcite{coavoux2017incremental} present a transition system which replaces \texttt{swap} with a \texttt{gap} transition. The intuition is that a reduction does not need to be always applied locally to the two topmost elements in the stack, and that those two items can be connected, despite the existence of a gap between them, using non-local reductions. Their algorithm ensures an upper-bound of $\frac{n(n-1)}{2}$ transitions.\footnote{Or alternatively $4n-2$, if we apply additional constraints to the \texttt{gap} transition and transitions following a \texttt{shift} action. However this comes at a cost of not being able to map more than $m$ gaps within the same discontinuous constituent.} With a different optimization goal, \newcite{stanojevic2017neural} removed the traditional reliance of discontinuous parsers on averaged perceptrons and hand-crafted features for a recursive neural network approach 
%(i.e. a Tree-\textsc{lstm}) 
that guides a \texttt{swap}-based system, with the capacity to generate contextualized representations. \newcite{coavoux2019discontinuous} replace the stack used in transition-based systems with a memory set containing the created constituents. This model allows interactions between elements that are not adjacent, without the \texttt{swap} transition, to create a new (discontinuous) constituent. Trained on a 2 stacked BiLSTM transducer, the model is guaranteed to build a tree with in 4$n$-2 transitions, given a sentence of length $n$.

A middle ground between explicit constituent parsing algorithms and this paper is the work based on transformations. For instance, \newcite{hall2008dependency} convert constituent trees into a non-linguistic dependency representation that is learned by a transition-based dependency parser, to then map its output back to a constituent tree. A similar approach is taken by \newcite{fernandez2015parsing}, but they proposed a more compact representation that leads to a much reduced set of output labels. Other authors such as \newcite{versley2016discontinuity} propose a two-step approach that approximates discontinuous structure trees by parsing context-free grammars with generative probabilistic models and transforming them to discontinuous ones.
\newcite{corro2017efficient} cast discontinuous phrase-structure parsing into a framework that jointly performs supertagging and non-projective dependency parsing by a reduction to the Generalized  Maximum Spanning  Arborescence problem \cite{myung1995generalized}. The recent work by \newcite{fernandez2020discontinuous} can be also framed within this paradigm. They essentially adapt the work by \newcite{fernandez2015parsing} and replace the averaged perceptron classifier with pointer networks \cite{vinyals2015pointer}, adressing the problem as a sequence-to-sequence task (for dependency parsing) whose output is then mapped back to the constituent tree. And next, \newcite{fernandez2020multitaskpointer} extended pointer networks with multitask learning to jointly predict constituent and dependency outputs.

In this context, the closest work to ours is the reduction proposed by \citet{gomez-rodriguez-vilares-2018-constituent}, who cast \emph{continuous} constituent parsing as sequence labeling.\footnote{Related to constituent parsing and sequence labeling, there are two related papers that made early efforts (although not a full reduction of the former to the latter) and need to be credited too. \newcite{ratnaparkhi1999learning} popularized maximum entropy models for parsing and combined a sequence labeling process that performs PoS-tagging and chunking with a set of shift-reduce-like operations to complete the constituent tree. In a related line, \newcite{collobert2011deep} proposed a multi-step approach consisting of $n$ passes over the input sentence, where each of them tags every word as being part of a constituent or not at one of the $n$ levels of the tree, using a IOBES scheme. } In the next sections we build on top of their work and: (i) analyze why their approach cannot handle discontinuous phrases, (ii) extend it to handle such phenomena, and (iii) train functional sequence labeling discontinuous parsers.

 \section{Preliminaries}
 
Let $\vec{w}=[w_0,w_1, ..., w_{|\vec{w}|-1}]$ be an input sequence of tokens, and $T_{|w|}$ the set of (continuous) constituent trees for sequences of length $|w|$; \newcite{gomez-rodriguez-vilares-2018-constituent} define an encoding function $\Phi: T_{|w|} \rightarrow L^{|w|}$ to map continuous constituent trees into a sequence of labels of the same length as the input. Each label, $l_i \in L$, is composed of three components $l_i=(n_i, x_i, u_i)$: 

\begin{itemize}
    \item $n_i$ encodes the number of levels in the tree in common between a word $w_i$ and $w_{i+1}$. To obtain a manageable output vocabulary space, $n_i$ is actually encoded as the difference $n_i - n_{i-1}$, with $n_{-1}=0$. We denote by $abs(n_i)$ the absolute number of levels represented by $n_i$. i.e. the total levels in common shared between a word and its next one.
    
    \item $x_i$ represents the lowest non-terminal symbol shared between $w_i$ and $w_{i+1}$ at level $abs(n_i)$.
    
    \item $u_i$ encodes a leaf unary chain, i.e. non-terminals that belong only to the path from the terminal $w_i$ to the root.\footnote{\emph{Intermediate} unary chains are compressed into a single non-terminal and treated as a regular branches.} Note that $\Phi$ cannot encode this information in ($n_i, x_i$), as these components always represent common information between $w_i$ and $w_{i+1}$.
\end{itemize}

\begin{figure}[hbtp]
\centering
\includegraphics[width=0.9\columnwidth]{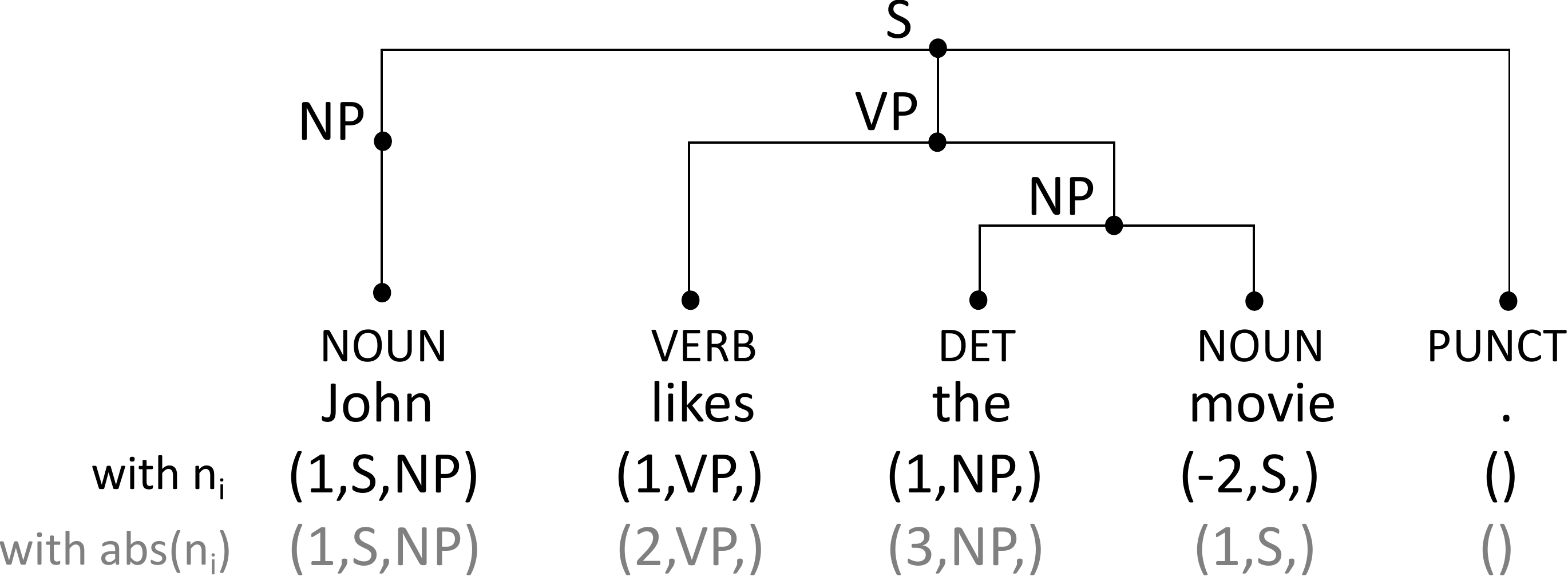}
\caption{\label{f-continuous-example} An example of a continuous tree encoded according to \newcite{gomez-rodriguez-vilares-2018-constituent}.}
\end{figure}

Figure \ref{f-continuous-example} illustrates the encoding on a continuous example.

\begin{figure}[hbtp]
\centering
\includegraphics[width=0.7\columnwidth]{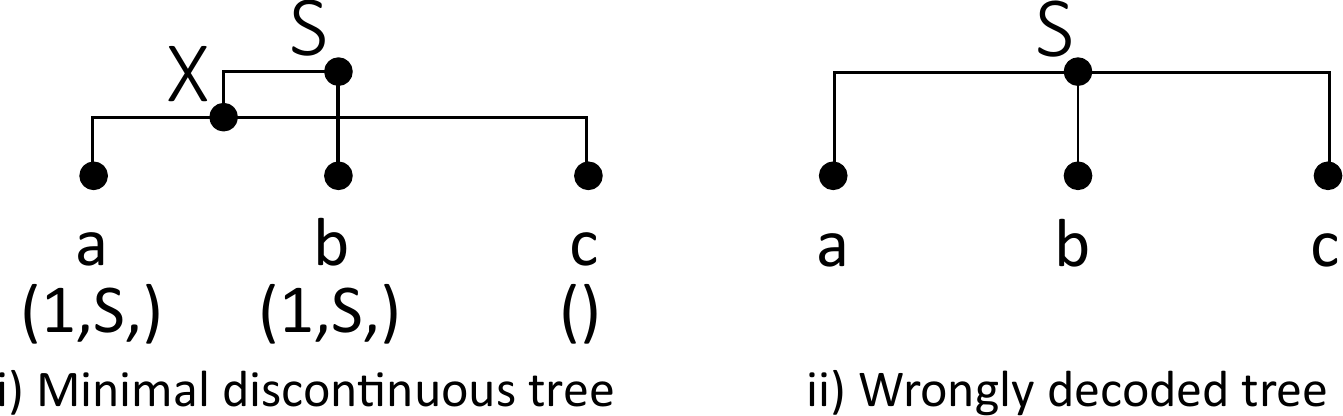}
\caption{\label{f-minimalist-discontinuous-tree} A minimal discontinuous constituent tree that cannot be decoded correctly if we rely on the \cite{gomez-rodriguez-vilares-2018-constituent} linearization.}
\end{figure}

\paragraph{Incompleteness for discontinuous phrase structures} \citeauthor{gomez-rodriguez-vilares-2018-constituent} proved that $\Phi$ is complete and injective for continuous trees. However, it is easy to prove that its validity does not extend to discontinuous trees, by using a counterexample. Figure \ref{f-minimalist-discontinuous-tree} shows a minimal discontinuous tree that cannot be correctly decoded. 

The inability to encode discontinuities lies on the assumption that $w_{i+1}$ will always be attached to a node belonging to the path from the root to $w_i$ ($n_i$ is then used to specify the location of that node in the path). This is always true in continuous trees, but not in discontinuous trees, as can be seen in Figure \ref{f-minimalist-discontinuous-tree} where $c$ is the child of a constituent that does not lie in the path from $S$ to $b$.

\section{Encoding nearly ordered permutations}\label{section-permutations}

\begin{figure*}[]
\centering
\includegraphics[width=2\columnwidth]{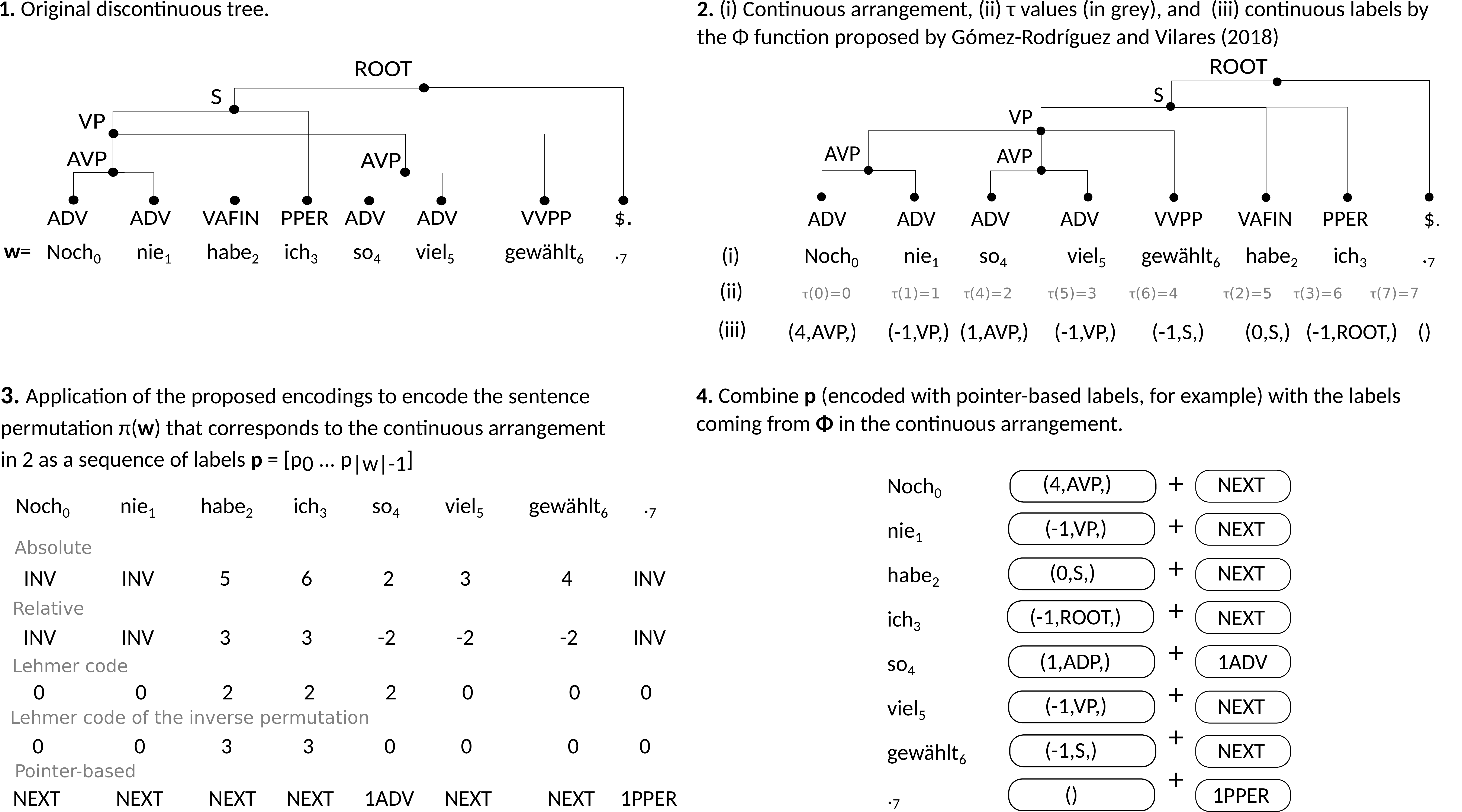}
\caption{\label{f-negra-discontinuous-tree-encoded} An example of the permutation encodings that allow for discontinuous parsing as sequence labeling}
\end{figure*}

Next, we fill this gap to address discontinuous parsing as sequence labeling. We will extend the encoding $\Phi$ to the set of discontinuous constituent trees, which we will call $T'_{|w|}$. The key to do this relies on a well-known property: a discontinuous tree $t \in T'_{|w|}$ can be represented as a continuous one using an in-order traversal that keeps track of the original indexes (e.g. the trees at the left and the right in Figure \ref{f-negra-discontinuous-tree-encoded}).\footnote{This is the discbracket format. See: \url{https://discodop.readthedocs.io/en/latest/fileformats.html}} We will call this tree the (canonical) continuous arrangement of $t$, $\omega(t) \in T_{|w|}$.

Thus, if given an input sentence we can generate  the position of every word as a terminal in  $\omega(t)$, the existing encodings to predict continuous trees as sequence labeling could be applied on $\omega(t)$. In essence, this is learning to predict a permutation of $\vec{w}$. As introduced in \S \ref{section-related-work}, the concept of location of a token is not a stranger in transition-based discontinuous parsing, where actions such as \texttt{swap} switch the position of two elements in order to create a discontinuous phrase. We instead propose to explore how to handle this problem in end-to-end sequence labeling fashion, without relying on any parsing structure nor a set of transitions.

To do so, first we denote by $\tau: \{0, \ldots, |w|-1\}  \rightarrow  \{0, \ldots, |w|-1\}$ the permutation that maps the position $i$ of a given $w_i$ in $\vec{w}$ into its position as a terminal node in $\omega(t)$.\footnote{Permutations are often defined as mappings from the element at a given position to the element that replaces it, but for our purpose, we believe that the definition as a function from original positions to rearranged positions (following, e.g., \citep{o2007discrete}) is more straightforward.}
From this, one can derive $\pi: W^n \rightarrow W^n$, a function that encodes a permutation of $\vec{w}$ in such way that its phrase structure does not have crossing branches. For continuous trees, $\tau$ and $\pi$ are identity permutations.
Then, we extend the tree encoding function $\Phi$ to $\Phi': T'_{|w|} \rightarrow L'^{|w|}$ where $l'_i \in L'$ is enriched with a fourth component $p_i$ such that $l'_i=(n_i,x_i,u_i,p_i)$, where $p_i$ is a discrete symbol such that the sequence of $p_i$'s encodes the permutation $\tau$ (typically each $p_i$ will be an encoding of $\tau(i)$, i.e. the position of $w_i$ in the continuous arrangement, although this need not be true in all encodings, as will be seen below).

The crux of defining a viable encoding for discontinuous parsing is then in how we encode $\tau$ as a sequence of values $p_i$, for $i=0 \ldots |\vec{w}|-1$. While the naive approach would be the identity encoding ($p_i = \tau(i)$), we ideally want an encoding that balances minimizing sparsity (by minimizing infrequently-used values) and maximizing learnability (by being predictable). To do so, we will look for encodings that take advantage of the fact 
that discontinuities in attested syntactic structures are mild \citep{maier:lichte:11}, i.e., in most cases, $\tau(i+1) = \tau(i)+1$. In other words, permutations $\tau$ corresponding to real syntactic trees tend to be nearly ordered permutations.
Based on these principles, we propose below a set of concrete encodings, which are also depicted on an example in Figure \ref{f-negra-discontinuous-tree-encoded}. All of them handle multiple gaps (a discontinuity inside a discontinuity) and cover 100\% of the discontinuities. Even if this has little effect in practice, it is an interesting property compared to algorithms that limit the number of gaps they can address \cite{coavoux2019discontinuous,corro2020span}.

\paragraph{Absolute-position} For every token $w_i$, $p_i = \tau(i)$ only if $i \neq \tau(i)$.
Otherwise, we use a special label \texttt{INV}, which represents that the word is a fixed point in the permutation, i.e., it occupies the same place in the sentence and in the continuous arrangement. 

\paragraph{Relative-position} If $i \neq \tau(i)$, then $p_i= i- \tau(i)$; otherwise, we again use the \texttt{INV} label.

\paragraph{Lehmer code}\cite{laisant1888numeration,lehmer1960teaching} In combinatorics, let $\vec{n}=[0,...,n-1]$ be a sorted sequence of objects, a Lehmer code is a sequence $\sigma=[\sigma_0,...\sigma_{n-1}]$ that encodes one of the $n!$ permutations of $\vec{n}$, namely $\vec{\alpha}$. The idea is intuitive: let $\vec{n^{i+1}}$ be the \emph{subsequence} of objects from $\vec{n}$ that remain available after we have permuted the first $i$ objects to achieve the permutation $\alpha$, then $\sigma_{i+1}$ equals the (zero-based) position in $\vec{n^{i+1}}$ of the next object to be selected. For instance, given $\vec{n}$ = $[0,1,2,3,4]$ and a valid permutation $\alpha$ = $[0,1,3,4,2]$, then $\vec{\sigma}$ = $[0,0,1,1,0]$. Note that the identity permutation would be encoded as a sequence of zeros.

In the context of discontinuous parsing and encoding $p_i$, $\vec{n}$ can be seen as the input sentence $\vec{w}$ where $\pi(\vec{w})$ is encoded by $\sigma$. The Lehmer code is particularly suitable for this task in terms of compression, as in most of the cases we expect (nearly) ordered permutations, which translates into the majority of elements of $\vec{\sigma}$ being zero.\footnote{For a continuous tree,  $\sigma_i =0\ \forall i \in [0,|w|-1]$.} However, this encoding poses some potential learnability problems. The root of the problem is that $\sigma_i$ does not necessarily encode $\tau(i)$, but $\tau(j)$ where $j$ is the index of the word that occupies the $i$th position in the continuous arrangement (i.e., $j = \tau^{-1}(i)$). In other words, this encoding is expressed following the order of words in the continuous arrangement rather than the input order, causing a non-straightforward mapping between input words and labels. For instance, in the previous example, $\sigma_2$ does not encode the location of the object $n_2$=2 but that of $n_3$=3.

\paragraph{Lehmer code of the inverse permutation} To ensure that each $p_i$ encodes $\tau(i)$, we instead interpret $p_i$ as meaning that $w_i$ should fill the $(p_i+1)$th currently remaining blank in a sequence $\sigma$ that is initialized as a sequence of blanks, i.e. $\sigma=[\circ,\circ,...,\circ]$. For instance, let $\vec{n}=[0,1,2,3,4]$ be the original input and $\pi(\vec{n})=[0,1,3,4,2]$ its desired continuous arrangement. 
At the first and second steps, $n_0=0$ and $n_1=1$ occupy the first available blanks (so $p_0=p_1=0$), generating partial arrangements of the form $[0,\circ,\circ,\circ,\circ]$ and $[0,1,\circ,\circ,\circ]$. Then, $n_2=2$ would need to fill the third empty blank (so $p_2=2$), and we obtain $[0,0,\circ,\circ,2]$. After that, $n_3$ and $n_4$ occupy the first available blank (so $p_3=p_4=0$). Thus, we obtain the desired arrangement $\sigma=[0,1,3,4,2]$, and the encoding is $[0,0,2,0,0]$. It is easy to check that this produces the Lehmer code for the inverse permutation to $\tau$. Hence, it shares the property that the identity permutation is encoded by a sequence of zeros, but it is more straightforward for our purposes as each $p_i$ encodes information about $\tau(i)$, the target position of $w_i$ in the continuous arrangement. Note that this and the Lehmer code coincide iff $\tau$ is a self-conjugate permutation (i.e., a conjugate that is its own inverse, see \citep{muir1891self}), of which the identity is a particular case.

\paragraph{Pointer-based encoding} When encoding $\tau(i)$, the previous encodings generate the position for the target word, but they do not really take into account the left-to-right order in which sentences are naturally read,\footnote{We use left-to-right in an informal sense
to mean that sentences are processed in linear temporal order. Of course, not all languages  follow a left-to-right script.}
nor they are linguistically inspired.

In particular, informally speaking, in human linguistic processing (i.e. when a sentence is read from left to right) we could say that a discontinuity is processed when we read a word that continues a phrase other than that of the previously read word.
For example, for the running example sentence (Figure \ref{f-negra-discontinuous-tree-encoded}), from an abstract standpoint we know that
there is a discontinuity because $\tau(2) \neq \tau(1)+1$, i.e., ``nie'' and ``habe'' are not contiguous in the continuous arrangement of the tree. However, in a left-to-right processing of the sentence, there is no way to know the final desired position of ``habe'' ($\tau(2)$) until we read the words ``so viel gew{\"a}hlt'', which go before it in the continuous arrangement. Thus, the requirement of the previous four encodings to assign a concrete non-default value to the $p_i$s associated with ``habe'' and ``ich'' is not too natural from an incremental reading standpoint, as learning $p_i$ requires information that can only be obtained by looking to the right of $w_i$. This can be avoided by using a model that just processes ``Noch nie habe ich'' as if it were a continuous subtree (in fact, if we removed ``so viel gew{\"a}hlt'' from the sentence, the tree would be continuous). Then, upon reading ``so'', the model notices that it continues the phrase associated with ``nie'' and not with ``ich'', and hence inserts it after ``nie'' in the continuous arrangement.

This idea of incremental left-to-right processing of discontinuities is abstracted in the form of a pointer $\widehat{o}$ that signals the last terminal in the
current continuous arrangement of the constituent that we are currently filling. That said, to generate the labels this approach needs to consider two situations:

\begin{itemize}

\item If $w_i$ is to be inserted right after $w_{i-1}$ (this situation is characterized by $\tau(i-1) < \tau(i) \wedge \nexists j<i \mid \tau(i-1) < \tau(j) < \tau(i)$).
This case is abstracted by a single label, $p_{i}$=\texttt{NEXT}, that means to insert at the position currently pointed by $\widehat{o}$, and then update $\widehat{o}=\tau^\prime_i(i)$, where the function $\tau^\prime_i$ is defined as $\tau^\prime_i(x) = \# \{ j \le i \mid \tau(j) \le \tau(x) \}$. $\tau_i'(x)$ can informally be described as a tentative value of $\tau(x)$, corresponding to the position of $w_x$ in the part of the continuous arrangement that involves the substring $w_0 \ldots w_i$.

\item Otherwise, $w_{i}$ should be inserted after some $w_{i-x}$
with $x \geq 1$, which means there is a discontinuity and that the current pointer $\widehat{o}$ is no longer valid and needs to be first updated to point to $\tau^\prime_i(i-x)$. To generate the label $p_{i}$ we use a tuple $(j,t)$ that indicates that the predecessor of $w_{i}$  in $\omega(t)$ is the $j$th preceding word in $\vec{w}$ with the PoS tag $t$. After that, we update the pointer to $\widehat{o}=\tau^\prime_i(i)$. 
While this encoding could work with PoS-tag-independent relative offsets, or any word property, the PoS-tag-based indexing provides linguistic grounding and is consistent with sequence labeling encodings that have obtained good results in dependency parsing \cite{strzyz2019viable}.

\end{itemize}

\paragraph{Pointer-based encoding (with simplified PoS tags)} A pointer-based variant where the PoS tags in $(j,t)$ are simplified (e.g. NNS $\rightarrow$ NN). The mapping is described in Appendix \ref{appendix-simplified-postags}. Apart from reducing sparsity, the idea is that a discontinuity is not so much influenced by specific information but by the coarse morphological category.\\

\noindent Ill-formed permutations are corrected with post-processing, following Appendix \ref{appendix-postprocessing}, to ensure that the derived permutations contain all word indexes.

\subsection{Limitations}

The encodings are complete under the assumption of an infinite label vocabulary. In practice, training sets are finite and this could cause the presence of unseen labels in the test set, especially for the integer-based label components:\footnote{This is a general limitation also present in previous parsing as sequence labeling approaches (e.g. \cite{gomez-rodriguez-vilares-2018-constituent}), and could potentially happen with any label component, e.g. predicting the non-terminal symbol. However, it is very unlikely that a non-terminal symbol has not been observed in the training set. Also, chart- and transition-based parsers would suffer from this same limitation.} the levels in common ($n_i$) and the label component $p_i$ that encodes $\tau(i)$. However, as illustrated in  Appendix \ref{appendix-unseen-labels-training}, an analysis on the corpora used in this work shows that the presence of unseen labels in the test set is virtually zero.

\section{Sequence labeling frameworks}\label{section-sequence-labeling-frameworks}

To test whether these encoding functions are learnable by parametrizable functions, we consider different sequence labeling architectures. We will be denoting by \textsc{encoder} a generic, contextualized encoder that for every word $w_i$  generates a hidden vector $\vec{h}_i$ conditioned on the sentence, i.e. \textsc{encoder($w_i|\vec{w}$)}=$\vec{h}_i$. We use a hard-sharing multi-task learning architecture \cite{caruana1997multitask,vilares2019better} to map every $\vec{h}_i$ to four 1-layered feed-forward networks, followed by softmaxes, that predict each of the components of $l'_i$. Each task's loss is optimized using categorical cross-entropy $\mathcal{L}_{t} = -\sum{log(P(l'_i|\vec{h}_i))}$ and the final loss computed as $\mathcal{L} = \sum_{t \in Tasks}{\mathcal{L}_{t}}$. We test four \textsc{encoder}s, which we briefly review but treat as black boxes. Their number of parameters and the training hyper-parameters are listed in Appendix \ref{appendix-hyperparameters}.

\paragraph{Transducers without pretraining} We try (i) a 2-stacked BiLSTM \cite{hochreiter1997long,yang2018ncrf++} where the generation of $\vec{h}_i$ is conditioned on the left and right context. 
(ii) We also explore a Transformer encoder \cite{vaswani2017attention} with 6 layers and 8 heads. The motivation is that we believe that the multi-head attention mechanism, in which a word attends to every other word in the sentence, together with positional embeddings, could be beneficial to detect discontinuities. In practice, we found training these transformer encoders harder than training BiLSTMs, and that obtaining a competitive performance required larger models, smaller learning rates, and more epochs (see also Appendix \ref{appendix-hyperparameters}).

The input to these two transducers is a sequence of vectors composed of: a pre-trained word embedding \cite{ling2015two} further fine-tuned during training, a PoStag embedding, and a second word embedding trained with a character LSTM. Additionally, the Transformer uses positional embeddings to be aware of the order of the sentence.

\paragraph{Transducers with pretraining} Previous work on sequence labeling parsing \cite{gomez-rodriguez-vilares-2018-constituent,strzyz2019viable} has shown that although effective, the models lag a bit behind state-of-the-art accuracy.
This setup, inspired in \newcite{vilares2020parsing}, aims to evaluate whether general purpose NLP architectures can achieve strong results when parsing free word order languages. In particular, we fine-tune (iii) pre-trained BERT \cite{devlin-etal-2019-bert}, and (iv) pre-trained DistilBERT \cite{sanh2019distilbert}. 
BERT and DistilBERT map input words to sub-word pieces \cite{wu2016google}. We align each word with its first sub-word, and use their embedding as the \emph{only}  input for these models. 

\section{Experiments}

\paragraph{Setup} For English, we use the discontinuous Penn Treebank (DPTB) by \newcite{evang2011plcfrs}. For German, we use TIGER and NEGRA \cite{brants2002tiger,skut1997annotation}. We use the splits by \newcite{coavoux2019discontinuous} which in turn follow the \newcite{dubey2003probabilistic} splits for the NEGRA treebank, the \newcite{seddah-etal-2013-overview} splits for TIGER, and the standard splits for (D)PTB (Sections 2 to 21
for training, 22 for development and 23 for testing). See also Appendix \ref{appendix-treebank-statistics} for more detailed statistics. We consider gold and predicted PoS tags. For the latter, the parsers are \emph{trained} on predicted PoS tags, which are generated by a 2-stacked BiLSTM, with the hyper-parameters used to train the parsers. The PoS tagging accuracy (\%) on the dev/test is: DPTB 97.5/97.7, TIGER 98.7/97.8 and NEGRA 98.6/98.1. BERT and DistilBERT do not use PoS tags as input, but when used to predict the pointer-based encodings, they are required to decode the labels into a parenthesized tree, causing variations in the performance.\footnote{The rest of BERT models do not require PoS tags at all.} Table \ref{table-treebank-label-stats} shows the number of labels per treebank.

\begin{table}[!phtb]
\centering
\small{
\tabcolsep=0.11cm
\begin{tabular}{l|rrr}
\hline
\multirow{2}{*}{Label Component} & \multicolumn{3}{c}{\#Labels}\\
& TIGER & NEGRA & DPTB \\
\hline
$n_i$&22&19&34\\
$x_i$&93&56&137\\
$u_i$&15&4&56\\
\hline
$p_i$ as absolute-position&129&110&98\\
$p_i$ as relative-position&105&90&87\\
$p_i$ as Lehmer&39&34&27\\
$p_i$ as inverse Lehmer&68&57&61\\
$p_i$ as pointer-based&122&99$^{\star}$&110$^{\star}$\\
$p_i$ as pointer-based simplified&81&65&83$^{\star}$\\
\hline
\end{tabular}}
\caption{\label{table-treebank-label-stats} Number of values per label component, merging the training and dev sets (gold setup). $\star$ are codes that generate one extra label with predicted PoS tags (this variability depends on the used PoS-tagger).
}
\end{table}

\begin{table*}[]
\centering
\small{
\begin{tabular}{lccccccc}
\hline
\multirow{2}{*}{Encoding} & \multirow{2}{*}{Transducer} & \multicolumn{2}{c}{TIGER} & \multicolumn{2}{c}{NEGRA} & \multicolumn{2}{c}{DPTB} \\
              & & F1 & Disco F-1 & F1 & Disco F-1 & F1 & Disco F-1 \\
\hline
Absolute-position&BiLSTM&75.2&12.4&72.8&12.8&86.0&10.7\\
Relative-position&BiLSTM&77.7&20.4&73.4&14.9&86.6&15.2\\
Lehmer&BiLSTM&81.6&33.4&76.8&26.2&88.4&30.7\\
Inverse Lehmer&BiLSTM&83.2&41.6&77.3&27.0&88.9&36.0\\
Pointer-based&BiLSTM&84.4&\bf 49.0&79.8&36.7&89.9&\bf 47.9\\
Pointer-based simplified&BiLSTM&\bf 84.6&48.7&\bf 79.8&\bf 38.1&\bf 90.0&46.3\\
\hline
Absolute-position&Transformer&81.9&38.3&75.3&25.4&87.5&25.8\\
Relative-position&Transformer&77.0&20.2&71.4&13.5&86.8&16.4\\
Lehmer&Transformer&82.6&38.5&75.4&21.4&88.1&24.8\\
Inverse Lehmer&Transformer&85.3&47.9&77.7&30.8&88.7&35.7\\
Pointer-based&Transformer&\bf 86.0&\bf51.2&79.8&38.8&90.2&\bf 46.7\\
Pointer-based simplified&Transformer&\bf 86.0&50.4&\bf80.6&\bf42.5&\bf90.2&46.2\\
\hline
Absolute-position&BERT&86.4&47.4&80.7&25.3&89.4&20.7\\
Relative-position&BERT&83.8&29.5&78.7&18.0&89.8&22.5\\
Lehmer&BERT&86.9&43.6&82.6&30.4&91.0&36.3\\
Inverse Lehmer&BERT&86.9&50.3&83.3&34.6&90.9&38.1\\
Pointer-based&BERT&\bf89.2&57.8&\bf 86.4&\bf 52.0&\bf 92.2&\bf 53.8\\
Pointer-based simplified&BERT&\bf 89.2&\bf 59.7&\bf86.4&49.3&92.0&50.9\\
\hline
Absolute-position&DistilBERT&82.0&30.6&75.6&19.0&88.2&17.7\\
Relative-position&DistilBERT&80.3&21.8&74.3&12.3&88.1&18.4\\
Lehmer&DistilBERT&83.3&32.8&77.6&21.6&89.5&33.0\\
Inverse Lehmer&DistilBERT&84.2&39.7&78.5&25.3&89.7&34.2\\
Pointer-based&DistilBERT&86.8&51.6&\bf 82.8&\bf 42.7&\bf 90.7&\bf 46.3\\
Pointer-based simplified&DistilBERT&\bf 87.0&\bf 54.7&82.7&40.5&\bf90.7&43.1\\

\hline
\end{tabular}}
\caption{\label{table-dev-results} Comparison of our approaches on the TIGER, NEGRA and DPTB dev splits (with gold PoS tags)}
\end{table*}

\paragraph{Metrics} We report the F-1 labeled bracketing score for all and discontinuous constituents, using \texttt{discodop}~\cite{vancranenburgh2016disc}\footnote{\url{http://github.com/andreasvc/disco-dop}} and the \texttt{proper.prm} parameter file. Model selection is based on overall bracketing F1- score.

\subsection{Results}

Table \ref{table-dev-results} shows the results on the dev sets for all encodings and transducers. The tendency is clear showing that the pointer-based encodings obtain the best results. The pointer-based encoding with simplified PoS tags does not lead however to clear improvements, suggesting that the models can learn the sparser original PoS tags set. For the rest of encodings we also observe interesting tendencies. For instance, when running experiments using stacked BiLSTMs, the relative encoding performs better than the absolute one, which was somehow expected as the encoding is less sparse. However, the tendency is the opposite for the Transformer encoders (including BERT and DistilBERT), especially for the case of discontinuous constituents. We hypothesize this is due to the capacity of Transformers to attend to every other word through multi-head attention, which might give an advantage to encode absolute positions over BiLSTMs, where the whole left and right context is represented by a single vector.  With respect to the Lehmer and  Lehmer of the inverse permutation encodings,
the latter performs better overall, confirming the bigger difficulties for the tested sequence labelers to learn Lehmer, which in some cases has a performance even close to the naive absolute-positional encoding (e.g. for TIGER using the vanilla Transformer encoder and BERT). As introduced in \S \ref{section-permutations}, we hypothesize this is caused by the non-straightforward mapping between words and labels (in the Lehmer code the label generated for a word does not necessarily contain information about the position of such word in the continuous arrangement).

In Table \ref{table-comparison-SOTA} we compare a selection of our models against previous work using both gold and predicted PoS tags. In particular, we include: (i) models using the pointer-based encoding, since they obtained the overall best performance on the dev sets, and (ii) a representative subset of encodings (the absolute positional one and the Lehmer code of the inverse permutation) trained with the best performing transducer. Additionally, for the case of the (English) DPTB, we also include experiments using a \texttt{bert-large} model, to shed more light on whether the size of the networks is playing a role when it comes to detect discontinuities. Additionally, we report speeds on CPU and GPU.\footnote{For CPU, we used a single core of an Intel(R) Core(TM) i7-7700 CPU @ 3.60GHz. For GPU experiments, we relied on a single GeForce GTX 1080, except for the BERT-large experiments, where due to memory requirements we required a Tesla P40.} The experiments show that the encodings are learnable, but that the model's power makes a difference. For instance, in the predicted setup BILSTMs and vanilla Transformers perform in line with pre-deep learning models \cite{maier2015discontinuous, fernandez2015parsing,coavoux2017incremental}, DistilBERT already achieves a robust performance, close to models such as \cite{coavoux2019discontinuous, coavoux2019unlexicalized}; and BERT transducers suffice to achieve results close to some of the strongest approaches, e.g. \cite{fernandez2020discontinuous}. Yet, the results lag behind the state of the art. With respect to the architectures that performed the best the main issue is that they are the bottleneck of the pipeline. Thus, the computation of the contextualized word vectors under current approaches greatly decreases the importance, when it comes to speed, of the chosen parsing paradigm used to generate the output trees (e.g. chart-based versus sequence labeling).

Finally, Table \ref{table-detailed-results-best-models} details the discontinuous performance of our best performing models.

\begin{table*}[!]
\centering
\small{
\tabcolsep=0.09cm
%\newcolumntype{g}{>{\columncolor{gray!10}}c}
\begin{tabular}{p{0.15cm}l|cccc|cccc|cccc}
\cline{2-14}
& \multirow{2}{*}{Model} & \multicolumn{4}{c|}{TIGER} & \multicolumn{4}{c|}{NEGRA} & \multicolumn{4}{c}{DPTB}\\
&   & F1 & Dis F-1 & CPU & GPU & F1 & Dis F-1 & CPU & GPU & F1 & Dis F-1 & CPU & GPU  \\
\cline{2-14}
\multirow{17}{*}{\STAB{\rotatebox[origin=l]{90}{\textcolor{gray}{Predicted PoS tags, own tags, or no tags}}}}
&\cg Pointer-based\tiny{ BiLSTM} & \cg 77.5&\cg 39.5&\cg 210&\cg \bf 568&\cg 75.6&\cg 34.6&\cg \bf 244&\cg \bf 715&\cg 88.8&\cg 45.8&\cg \bf 194&\cg \bf 611\\
&\cg Pointer-based\tiny{ Transformer} &\cg 78.3&\cg 41.2&\cg 97&\cg 516&\cg 75.0&\cg 33.6&\cg 118&\cg 659&\cg 89.3&\cg 45.2&\cg 104&\cg 572\\
&\cg Pointer-based\tiny{ DistillBERT} &\cg 81.3&\cg 43.2&\cg 5&\cg 145&\cg 81.0&\cg 41.5&\cg 5&\cg 147&\cg 90.1&\cg 41.0&\cg 5&\cg 142\\
&\cg Pointer-based\tiny{ BERT base} &\cg 84.6&\cg 51.1&\cg 2&\cg 80&\cg 83.9&\cg 45.6&\cg 2&\cg 80&\cg 91.9&\cg 50.8&\cg 2&\cg 80\\
&\cg Pointer-based\tiny{ BERT large} &\cg -&\cg -&\cg -&\cg -&\cg -&\cg -&\cg -&\cg -&\cg 92.8 &\cg 53.9&\cg 0.75&\cg 34\\
&\cg Absolute-position\tiny{ BERT base}&\cg 80.3& \cg 34.8&\cg 2&\cg 80&\cg 76.6&\cg 22.6&\cg 2&\cg 81&\cg 88.8&\cg 18.0&\cg 2&\cg 79\\
&\cg Inverse Lehmer\tiny{ BERT base}&\cg 81.5&\cg 38.7&\cg 2&\cg 80&\cg 80.5&\cg 34.4&\cg 2&\cg 81&\cg 89.7&\cg 29.3&\cg 2&\cg 80\\
&
\defcitealias{fernandez2020multitaskpointer}{Fern\'andez-G. and G\'omez-R. (2020b)}\citetalias{fernandez2020multitaskpointer}
&86.6&62.6&-&-&86.8&69.5&-&-&-&-&-&-\\
&\defcitealias{fernandez2020multitaskpointer}{Fern\'andez-G. and G\'omez-R. (2020b)}\citetalias{fernandez2020multitaskpointer}$^\star$&89.8&\bf 71.0&-&-&91.0&\bf 76.6&-&-&-&-&-&-\\
&\newcite{stanojevic2020span}&83.4&53.5&-&-&83.6&50.7&-&-&90.5&67.1&-&-\\
&\newcite{corro2020span}\tiny{ $\mathcal{O}(n^3)$}&85.2&51.2&-&474&86.3&56.1&-&478&92.9&64.9&-&355\\
&\newcite{corro2020span}\tiny{ $\mathcal{O}(n^6)$}&84.9&51.0&-&3&85.6&53.0&-&41&92.6&59.7&-&22\\
&\newcite{corro2020span}\tiny{ $\mathcal{O}(n^3)$}\small{$^\star$}&\bf 90.0& 62.1&-&-&\bf 91.6&66.1&-&-&\bf 94.8&\bf 68.9&-&-\\
&\defcitealias{fernandez2020discontinuous}{Fern\'andez-G. and G\'omez-R. (2020a)}\citetalias{fernandez2020discontinuous}&85.7&60.4&-&-&85.7&58.6&-&-&-&-&-&-\\
&\newcite{coavoux2019discontinuous}$^\diamond$&82.5&55.9&64&-&83.2&56.3&-&-&90.9&67.3&38&-\\
&\newcite{coavoux2019unlexicalized}$^\diamond$&82.7&55.9&126&-&83.2&54.6&-&-&91.0&71.3&80&-\\
&\newcite{coavoux2017incremental}$^\diamond$&79.3&-&\bf 260&-&-&-&-&-&-&-&-&-\\
&\newcite{corro2017efficient}&-&-&-&-&-&-&-&-&89.2&-&7&-\\
&\newcite{stanojevic2017neural}&77.0&-&-&-&-&-&-&-&-&-&-&-\\
&\newcite{versley2016discontinuity}&79.5&-&-&-&-&-&-&-&-&-&-&-\\
&\defcitealias{fernandez2015parsing}{Fern\'andez and Martins (2015)}\citetalias{fernandez2015parsing}&77.3&-&-&-&77.0&-&37&-&-&-&-&-\\

\cline{2-14}

\multirow{15}{*}{\STAB{\rotatebox[origin=l]{90}{\textcolor{gray}{Gold PoS tags}}}}
&\cg Pointer-based\tiny{ BiLSTM} &\cg 79.2 &\cg 40.1 &\cg 210&\cg \bf 568&\cg 77.1 &\cg 36.5 &\cg \bf 244&\cg \bf 715&\cg 89.1 &\cg 41.8&\cg \bf 194&\cg \bf 611 \\
&\cg Pointer-based\tiny{ Transformer} &\cg 79.4 &\cg 41.0 &\cg 97&\cg 516&\cg 77.1 &\cg 34.9 &\cg 118&\cg 659&\cg 89.9 &\cg 48.0&\cg 104&\cg 572 \\
&\cg Pointer-based\tiny{ DistillBERT} &\cg 81.4 &\cg 43.8 &\cg 5&\cg 145&\cg 80.7 &\cg 36.8 &\cg 5&\cg 147&\cg 90.4 &\cg 42.7&\cg 5&\cg 142 \\
&\cg Pointer-based\tiny{ BERT base} &\cg 84.7 &\cg 51.6 &\cg 2&\cg 80&\cg 84.2 &\cg 46.9 &\cg 2&\cg 81&\cg 91.7 &\cg 49.1&\cg 2&\cg 80 \\
&\cg Pointer-based\tiny{ BERT large} &\cg -&\cg -&\cg - &\cg - &\cg -  &\cg -  &\cg -&\cg -&\cg \bf 92.8&\cg \bf 55.4&\cg 0.75&\cg 34\\
&\defcitealias{fernandez2020multitaskpointer}{Fern\'andez-G. and G\'omez-R. (2020b)}\citetalias{fernandez2020multitaskpointer}&\bf 87.3&\bf 64.2&-&-&\bf 87.3&\bf 71.0&-&-&-&-&-&-\\
&\defcitealias{fernandez2020discontinuous}{Fern\'andez-G. and G\'omez-R. (2020a)}\citetalias{fernandez2020discontinuous}& 86.3& 60.7&-&-& 86.1& 59.9&-&-&-&-&-&-\\
&\newcite{coavoux2017incremental}$^\diamond$&81.6&49.2&\bf 260&-&82.2&50.0&-&-&-&-&-&-\\
&\newcite{corro2017efficient}&81.6&-&-&-&-&-&-&-&90.1&-&7&-\\
&\newcite{stanojevic2017neural} &81.6&-&-&-&82.9&-&-&-&-&-&-&-\\
&\newcite{maier2016discontinuous} &76.5&-&-&-&-&-&-&-&-&-&-&-\\
&\defcitealias{fernandez2015parsing}{Fern\'andez-G. and Martins (2015)}\citetalias{fernandez2015parsing} &80.6&-&-&-&80.5&-&37&-&-&-&-&-\\
&\newcite{maier2015discontinuous}\tiny{ beam search}$^\diamond$&74.7&18.8&73&&77.0&19.8&80&-&-&-&-&-\\
&\newcite{maier2015discontinuous}\tiny{ greedy}$^\diamond$&-&-&-&-&-&-&640&-&-&-&-&-\\

\cline{2-14}

\end{tabular}}
\caption{\label{table-comparison-SOTA} Comparison against related work on the TIGER, NEGRA and DPTB test splits. The $\star$ symbol indicates that a model used BERT to contextualize the input. The reported speeds are extracted from the related work and therefore results are not directly comparable since the hardware can be different. The $\diamond$ symbol indicates work that reported the speed (in sentences per second) on the dev sets instead.}
\end{table*}

\begin{table*}[!hbtp]
\begin{center}
\small
\begin{tabular}{l|rrr|rrr|rrr}
\hline
 \multirow{2}{*}{Model} & \multicolumn{3}{c|}{TIGER} & \multicolumn{3}{c|}{NEGRA} & \multicolumn{3}{c}{DPTB} \\
 & Dis P & Dis R & Dis F-1 & Dis P & Dis R & Dis F-1 & Dis P & Dis R & Dis F-1 \\
 \hline
 Pointer-based \tiny{BiLSTM}& 41.0 & 38.1 & 39.5 & 34.7 & 34.5 & 34.6 & 46.7 & 45.0 & 45.8 \\
 Pointer-based \tiny{Transformer} & 39.0 & 43.8 & 41.2 & 30.4 & 37.7 & 33.6 & 43.3 & 47.2 & 45.2 \\
  Pointer-based \tiny{DistillBERT} & 42.5 & 43.9 & 43.2 & 41.0 & 42.0 & 41.5 & 37.6 & 45.0 & 41.0 \\
 Pointer-based \tiny{BERT} &\bf 50.9 &\bf 51.4 &\bf 51.1 &\bf 43.2 &\bf 48.4 &\bf 45.6 &\bf 47.9 &\bf 54.0 &\bf 50.8 \\

 \hline
 
\end{tabular}
\end{center}
\caption{\label{table-detailed-results-best-models} Detailed discontinuous performance (Discontinuous Precision, Recall and F1-score) by our best sequence labeling models (predicted PoS tags setup).}
\end{table*}

\paragraph{Discussion on other applications} It is worth noting that while we focused on parsing as sequence labeling, 
encoding syntactic trees as labels is useful to straightforwardly feed syntactic information to downstream models, even if the trees themselves come from a non-sequence-labeling parser. For example, \citet{wang-etal-2019-best} use the sequence labeling encoding of \citet{gomez-rodriguez-vilares-2018-constituent} to provide syntactic information to a semantic role labeling model. Apart from providing fast and accurate parsers, our encodings can be used to do the same with discontinuous syntax.

\section{Conclusion}

We reduced discontinuous parsing to sequence labeling. The key contribution consisted in predicting a continuous tree with a rearrangement of the leaf nodes to shape discontinuities, and defining various ways to encode such a rearrangement as a sequence of labels associated to each word, taking advantage of the fact that in practice they are nearly ordered permutations.
We tested whether those encodings are learnable by neural models and saw that the choice of permutation encoding is not trivial, and there are interactions between encodings and models (i.e., a given architecture may be better at learning a given encoding than another). Overall, the models achieve a good trade-off speed/accuracy without the need of any parsing algorithm or auxiliary structures, while being easily parallelizable.

\section*{Acknowledgments}

We thank Maximin Coavoux for giving us access to the data used in this work. We acknowledge the European Research Council (ERC), which has funded this research under the European Union's Horizon 2020 research and innovation programme (FASTPARSE, grant agreement No 714150), MINECO (ANSWER-ASAP, TIN2017-85160-C2-1-R), Xunta de Galicia (ED431C 2020/11), and Centro de Investigación de Galicia "CITIC", funded by Xunta de Galicia and the European Union (European Regional Development Fund- Galicia 2014-2020 Program), by grant ED431G 2019/01.

\bibliography{disco}
\bibliographystyle{acl_natbib}

\clearpage
\appendix

\section{Appendices}\label{section-appendix}

\subsection{Simplified part-of-speech tags for the pointer-based encoding}\label{appendix-simplified-postags}

Table \ref{table-dptb-simplified-tags} maps the original PoS tags in the DPTB treebank into the simplified ones used for the second variant of the pointer-based encoding. Table \ref{table-tiger-negra-simplified-tags} does the same but for the TIGER and NEGRA treebanks.

\begin{table}[hbtp]
\begin{center}
\small
\begin{tabular}{lr}
\hline
  Original label &  Coarse label \\
 \hline
CC & CC\\
CD & CD\\
DT & DT\\
EX & EX\\
FW & FW\\
IN & IN\\
JJ,JJR,JJS & JJ\\
LS & LS\\
MD & MD\\
NN,NNS,NNP,NNPS & NN\\
PDT & PDT\\
POS & POS\\

PRP,PRP\$ & PRP\\
RB,RBR,RBS & RB\\
RP & RP\\
SYM & SYM\\
TO & TO\\
UH & UH\\
VB,VBD,VBG,VBN,VBP,VBZ & V\\
WDT,WP,WP\$,WRB & W\\

 \hline

\hline

\end{tabular}
\end{center}
\caption{\label{table-dptb-simplified-tags} Mapping from the original labels to coarse labels in the DPTB treebank}
\end{table}

\begin{table}[bpth]
\begin{center}
\small
\begin{tabular}{lr}
\hline
 Original label &  Coarse label \\
 \hline
NN,NE & N\\
ADJA,ADJD & ADJ\\
CARD & CARD\\
VAFIN,VAIMP,VVFIN,VVIMP,VMFIN & V\\
VVINF,VAINF,VMINF,VVIZU & V\\
VVPP,VMPP,VAPP & V\\
ART & ART\\
PPER,PRF,PPOSAT,PPOSS,PDAT,PDS & P\\
PIDAT,PIS,PIAT,PRELAT,PRELS & P\\
PWAT,PWS,PWAV,PAV & P\\
ADV,ADJD & AD\\
KOUI,KOUS,KON,KOKOM & K\\
APPR,APPRART,APPO,APZR & AP\\
PTKZU,PTKNEG,PTKVZ& PT\\
,PTKA,PTKANT & PT \\
\$,\$(,\$. & \$ \\
%#Sonstige (ITJ, TRUNC, XY, FM) There is no a clear pattern so we do not normalize them
ITJ & ITJ \\
TRUNC & TRUNC \\
XY & XY \\
FM & FM \\
 \hline

\end{tabular}
\end{center}
\caption{\label{table-tiger-negra-simplified-tags} Mapping from the original labels to coarse labels in the TIGER and NEGRA treebanks}
\end{table}

\subsection{Postprocessing of corrupted outputs}\label{appendix-postprocessing}

We describe below the post-processing of the encodings to ensure that the generated sequences can be later decoded to a well-formed tree. Before post-processing the predicted permutation, we make sure that one, and only one label $(n_i,x_i,u_i, p_i)$, can be identified as the last word in the continuous arrangement. This is required because the component $n_i$ encodes unique information for the last word (an empty dummy value, as $n_i$ always encodes information between a word and the next one, which does not exist for the last token); which can conflict with some of the predicted $p_i$s, that might put a different word into the last position. That said, we rely on the value $n_i$ to identify which word should be located as the last one.\footnote{If more than one $n_i$ refers to the last word, we consider the one with the largest index.}

\paragraph{Absolute-position and relative-position encodings} Given the sequence $\vec{p}$ that encodes the permutation $\pi(\vec{w})$ of the words of $\vec{w}$ in the continuous arrangement $\omega(t)$, we: (i) fill the indexes for which the predicted labels indicate that the token should remain in the same position, i.e. $p_i$=\texttt{INV}, and (ii) for the remaining $p_i$'s we check whether the predicted index has not been yet filled, and otherwise assign it to the closest available index (computed as the minimum absolute difference).

\paragraph{Lehmer encoding} Given $\vec{p}$ and the list of available word indexes \texttt{idxs} (initially all the words), we process the elements in $\vec{p}$ in a left-to-right fashion: (i) if the corresponding index encoded at $p_i$ is in \texttt{idxs}, then we select the index and remove it from \texttt{idxs}, (ii) otherwise, we select the last element in \texttt{idxs} and, again, remove it.

\paragraph{Lehmer of the inverse permutation encoding} The post-processing is similar to the Lehmer code encoding, but considering the available blanks instead of a list of word indexes.

\paragraph{Pointer-based encodings} Given the encoded permutation $\vec{p}$, we process the elements left-to-right and: (i) if $p_i$=\texttt{NEXT}, then we apply no post-processing and we consider the word will be inserted after the current pointer $\widehat{o}$ at the moment of decoding, which is always valid. (ii) Otherwise, we are processing an element $p_i$ that encodes the pointer $\widehat{o}=(j,t)$, and try to map it to $\tau^\prime(i)$. If such mapping is not possible, this is because $j$ is greater than the number of previously processed words that have the postag $t$. If so, then we post-process $p_i$ to $(k,t)$, where $k$ was the first processed word with postag $t$, or to $p_i$=\texttt{NEXT}, if there is no previous word labeled with the postag $t$.  

\subsection{Unseen labels in the training set}\label{appendix-unseen-labels-training}

Table \ref{table-missing-labels-in-test} shows some statistics on the number of gold $p_i$ label components that are present in the test sets, but not on the corresponding training or dev splits. We do not show statistics for the label component that represents the number of levels in common ($n_i$) since for all treebanks the number of missing $n_i$ values on the test sets was zero. For $p_i$, the missing elements correspond to rare situations. For instance, taking NEGRA as our reference corpus,\footnote{We consider NEGRA as the reference corpus since it was the treebank that showed the largest percentage of missing $p_i$ elements.} for the relative index encoding the only missing element was `-29' and occurred a single time in the test set; and for the pointer-based encoding the missing components were just three: `-10~NN' (2 occurrences), `-4~ADV' (1 occurrence), `-4~\$[' (1 occurrence).

\begin{table}[hbtp]
\begin{center}
\tabcolsep=0.13cm
\small
\begin{tabular}{p{2.3cm}l|ccc}
\hline
 \multirow{2}{*}{Encoding} & \multirow{2}{*}{Treebank} &\multicolumn{3}{c}{Missing elements}\\
& & Unique & Total & \%    \\
 \hline
   \multirow{3}{*}{Absolute} & TIGER &   6 & 6 & 6.5$\times 10^{-3}$ \\
                               & NEGRA &  0& 0 &0 \\
                               & DPTB  & 0& 0 & 0 \\ 
  \multirow{3}{*}{Relative} & TIGER  & 1 & 8 & 8.7$\times 10^{-3}$ \\
                               & NEGRA & 1 &1 & 5.9$\times 10^{-3}$  \\
                               & DPTB  & 0 & 0 & 0 \\ 
 \multirow{3}{*}{Lehmer} & TIGER & 0 &0 & 0\\
                               & NEGRA & 1 &1& 5.9$\times 10^{-3}$  \\
                               & DPTB  & 0 &0& 0 \\ 
 \multirow{3}{*}{Inverse Lehmer} & TIGER & 1 &1& 1.1$\times 10^{-3}$ \\
                               & NEGRA & 0 &0& 0 \\
                               & DPTB  & 0 &0& 0 \\ 
 \multirow{3}{*}{Pointer-based} & TIGER & 1 &1& 1.1$\times 10^{-3}$ \\
                               & NEGRA & 3 &4& 0.02\\
                               & DPTB  & 0 &0&0  \\ 
\multirow{3}{*}{Pointer-based simp.} & TIGER  & 1 &1& 1.1$\times 10^{-3}$ \\
                               & NEGRA & 1 &1& 5.9$\times 10^{-3}$\\
                               & DPTB  & 0 &0& 0 \\ 
 \hline

 \hline
 
\end{tabular}
\end{center}
\caption{\label{table-missing-labels-in-test} Number of unique $p_i$ label components that occur on the test set but not on the training or dev splits, total ocurrences and the corresponding percentage over the total number of labels.}
\end{table}

\subsection{Training hyper-parameters and size of the trained models}\label{appendix-hyperparameters}

Table \ref{table-hyper-bilstm-transducer} shows the hyper-parameters used to train the  BiLSTMs, both for the gold and predicted setups. We use pre-trained embeddings for English and German \cite{ling2015two}. The embeddings for English have 100 dimensions, while the German ones only have 60. For the BiLSTMs, we did not do any hyper-parameter engineering and just considered the hyper-parameters reported by \newcite{gomez-rodriguez-vilares-2018-constituent}.

\begin{table}[hbtp]
\begin{center}
\small
\begin{tabular}{lr}
\hline
 Hyperparameter &  Value \\
 \hline
 BiLSTM size&800 \\
 \# BiLSTM layers& 2\\
 optimizer& SGD\\
 loss& cat. cross-entropy\\
 learning rate&0.2\\
 decay (linear)&0.05\\
 momentum&0.9\\
 dropout&0.5\\
 word embs& \newcite{ling2015two}\\
 PoS tags emb size&20\\
 character emb size&30\\
 batch size training&8 \\
 training epochs&100\\
 batch size test&128 \\
 \hline
 
\hline

\end{tabular}
\end{center}
\caption{\label{table-hyper-bilstm-transducer} Main hyper-parameters for the training of the BiLSTMs, both for the gold and predicted setups}
\end{table}

Table \ref{table-hyper-transformer-transducer} shows the configuration used to train the vanilla Transformer encoders. As explained in the paper, we found out that Transformers were more unstable during training in comparison to BILSTMs. To overcome such unstability, we performed a small manual hyper-parameter search. This translated into training during more epochs, with a low learning rate and large dropout. 

\begin{table}[!]
\begin{center}
\small
\tabcolsep=0.06cm
\begin{tabular}{lcc}
\hline
  Hyperparameter &  Value &  Value \\
 &(gold setup)&(pred setup)\\
 \hline
 Att. heads & 8 & 8 \\
 Att. layers & 6 & 6 \\
 Hidden size & 800 & 800 \\
 Hidden dropout &0.4 & 0.4\\
 optimizer&SGD&SGD\\
 loss&cross-entropy&cross-entropy\\
 learning rate&0.004$^\star$&0.003\\
 decay (linear)&0.0&0.0\\
 momentum&0.0&0.0\\
 word embs& \newcite{ling2015two} & \newcite{ling2015two}\\
 PoS tags emb size&20&20\\
 character emb size&136/132$^\triangle$&136/132$^\triangle$\\
 batch size training&8&8 \\
 training epochs&400&400\\
 batch size test&128&128 \\
 \hline
 
\hline

\end{tabular}
\end{center}
\caption{\label{table-hyper-transformer-transducer}Main hyper-parameters for the training of the vanilla Transformer encoder, both for the gold and predicted setups. $\star$ Except for the pointer-based encoding, where 0.003 was necessary to converge. $\triangle$ The character embedding size used for the TIGER and NEGRA models, so the size of the input to the model is a multiplier of the number of attention heads. As \newcite{ling2015two} embeddings for German only have 60 dimensions, this tweak was necessary for those treebanks.}
\end{table}

To fine-tune the BERT and DistilBERT models we use the default fine-tuning setup provided by huggingface.\footnote{See \url{https://github.com/huggingface/transformers} and \url{https://github.com/aghie/disco2labels/blob/master/run_token_classifier.py}} Table \ref{table-hyper-bert-models} shows the  hyper-parameters that we have modified. For English, the pre-trained models we relied on were: `bert-base-cased', `distilbert-base-cased' (distilled from `bert-base-cased'), and `bert-large-cased'. For German, we used `bert-base-german-dbmdz-cased' and `distilbert-base-german-cased' (distilled from `bert-base-german-dbmdz-cased').

\begin{table}[!]
\begin{center}
\small
\begin{tabular}{lr}
\hline
 Hyperparameter &  Value \\
 \hline
 loss& cross-entropy\\
 learning rate&$1e^{-5}$\\
 batch size training&6 \\
 training epochs&45$^\star$\\
 batch size test&8 \\
 \hline
 
\hline

\end{tabular}
\end{center}
\caption{\label{table-hyper-bert-models} Main hyper-parameters for the training of the BERT and DistilBERT models, both for the gold and predicted setups. $\star$ except for BERT large, where we trained for 30 epochs.}
\end{table}

Finally, in Table \ref{table-number-hyperparameters} we list the number of parameters for each of the transducers trained on the pointer-based encoding. For the rest of the encodings, the models have a similar number of parameters, as the only change in the architecture is the small part involving the feed-forward output layer that predicts the label component $p_i$. 

\begin{table}[!hbtp]
\begin{center}
\small
\begin{tabular}{lrrr}
\hline
 Transducer & TIGER & NEGRA & DPTB \\
 \hline
 BiSLTM & 11.1M & 8.9M & 10.5M \\
 Transformer & 8.6M & 6.5M & 8.8M \\
 DistilBERT & 67.0M & 67.0M & 65.5M \\
 BERT-base & 110.1M & 110.1M &108.6M \\
 BERT-large &  &  & 333.9M \\
 \hline
 
\hline

\end{tabular}
\end{center}
\caption{\label{table-number-hyperparameters} Number of parameters (millions) for each model with the pointer-based encoding.}
\end{table}

More in detail, for BiLSTMs and vanilla Transformers the TIGER model is the larger than the NEGRA one. This is because for these transducers we only store and use the word embeddings from \newcite{ling2015two} that were seen in the training and dev sets, and the TIGER treebank is larger and contains more unique words. Also, we see that for the BiLSTMs the TIGER model is slightly larger than the DPTB one, while for the vanilla Transformer the opposite happens. This is due to the smaller char embedding size in the case of the German Transformers, which is required so the total size of the input vector is divisible by 8, the number of attention heads (the root of the need for the disparity in the char embedding sizes is that the pre-trained English and German embeddings also have a different number of dimensions). On the contrary, for the BERT-based models we use the same pre-trained model for TIGER and NEGRA, for example, which causes these models to have an almost identical number of parameters.

\subsection{Treebank statistics}\label{appendix-treebank-statistics}

Table \ref{table-splits} shows the number of samples per treebank split.
\begin{table}[hbtp]
\begin{center}
\small
\begin{tabular}{lrrr}
\hline
 Treebank &Training & Dev & Test \\
 \hline
 TIGER& 40\,472 & 5\,000 & 5\,000 \\
 NEGRA& 18\,602 & 1000 & 1000 \\
 DPTB & 39\,832 & 1\,700 & 2\,416  \\
 \hline
\end{tabular}
\end{center}
\caption{\label{table-splits} Number of samples per treebank split.}
\end{table}

\end{document}